\pdfoutput=1

\documentclass[11pt,a4paper]{article}
\usepackage[hyperref]{eacl2021}
\usepackage{times}
\usepackage{latexsym}

\usepackage{amsmath}
\usepackage{amssymb}
\usepackage{graphicx}
\usepackage{multirow}

\usepackage{xcolor}

\usepackage{microtype}

\usepackage{booktabs}
\usepackage{multirow}

\aclfinalcopy %

\title{Segmenting Subtitles for Correcting ASR Segmentation Errors}

\author{David Wan,$^{1}$
    Chris Kedzie,$^{1}$
    Faisal Ladhak,$^{1}$
    Elsbeth Turcan,$^{1}$ \\
    {\bf \large Petra Galu\v{s}\v{c}\'{a}kov\'{a},$^{2}$
    Elena Zotkina,$^{2}$
    Zhengping Jiang,$^{1}$
    }\\
    {\bf \large
    Peter Bell ${^3}$
    and Kathleen McKeown$^{1}$
    }
\\
  $^{1}$ Columbia University,
  $^{2}$ University of Maryland,
  $^{3}$ University of Edinburgh\\
  \{dw2735, zj2265\}@columbia.edu, 
  \{kedzie, faisal, eturcan, kathy\}@cs.columbia.edu,\\
  \{petra, elena\}@umiacs.umd.edu, 
  peter.bell@ed.ac.uk\\
  }

\date{}

\begin{document}
\maketitle
\begin{abstract}
Typical ASR systems segment the input audio into \textit{utterances} using purely acoustic information, which may not resemble the sentence-like units that are expected by conventional machine translation (MT) systems for Spoken Language Translation. 
In this work, we propose a model for correcting the 
acoustic segmentation of  ASR models
for low-resource languages
to improve performance 
on downstream tasks.
We propose the use of subtitles as a proxy dataset for correcting ASR acoustic segmentation,
creating synthetic acoustic utterances by modeling common error modes. 
We train a neural tagging model for correcting ASR acoustic segmentation and show that it improves downstream performance on MT and audio-document cross-language information retrieval (CLIR). 

\end{abstract}

\newcommand{\tokens}{\mathbf{x}}
\newcommand{\token}{x}
\newcommand{\ctags}{\mathbf{y}}
\newcommand{\ctag}{y}
\newcommand{\numToks}{n}
\newcommand{\atags}{\boldsymbol{\gamma}}
\newcommand{\atag}{\gamma}
\newcommand{\tokVocab}{\mathcal{V}}
\newcommand{\reals}[1]{\mathbb{R}^{#1}}
\newcommand{\binspace}[1]{\{0,1\}^{#1}}
\newcommand{\model}{p}
\newcommand{\params}{\theta}
\newcommand{\usegprob}{\check{\alpha}}
\newcommand{\utags}{\boldsymbol{\check{\atags}}}
\newcommand{\otags}{\boldsymbol{\hat{\atags}}}
\newcommand{\osegprob}{\hat{\alpha}}
\newcommand{\bernoulli}[1]{\operatorname{Bernoulli}(#1)}
\newcommand{\numSegs}{m}
\newcommand{\joinprob}{\kappa}
\newcommand{\emb}{\mathbf{e}}
\newcommand{\hid}{\mathbf{h}}
\newcommand{\corpus}{\mathcal{D}}
\newcommand{\lstm}{\operatorname{LSTM}}
\newcommand{\flstm}{\overrightarrow{\lstm}}
\newcommand{\blstm}{\overleftarrow{\lstm}}

\section{Introduction}

Typical ASR systems segment the input audio into \textit{utterances} using purely acoustic information, i.e., pauses in speaking or other dips in the audio signal, which may not resemble the sentence-like units that are expected by conventional MT systems for spoken language translation (SLT) \cite{segmentationimportance}.
Longer utterances may span multiple sentences, while shorter utterances may be sentence fragments containing only a few words (see \autoref{fig:newexample} for examples).
Both can be problematic for downstream MT systems. In this work, we propose a model for correcting the 
acoustic segmentation of an ASR model to improve performance 
on downstream tasks,
focusing on the challenges inherent to SLT pipelines for low-resource
languages.

While prior work has trained intermediate components to segment 
ASR output into sentence-like units
\cite{Matusov07improvingspeech,rao2007optimizing}, these have 
primarily focused on highly resourced language pairs such as Arabic and Chinese. 
When the source language is low-resource, suitable training data may be very limited for ASR and MT, and even nonexistent for segmentation. Since typical low-resource language ASR datasets crawled from the web do not have hand-annotated segments, we propose deriving proxy segmentation datasets from film and television subtitles.
Subtitles typically contain segment boundary information like sentence-final punctuation,
and while they are not exact transcriptions,
they are closer to transcribed speech than many other large text corpora.

Our proposed model takes as input a sequence of tokens and segmentation boundaries produced by the acoustic segmentation of the 
ASR system and returns a corrected segmentation.
While subtitles are often similar to speech transcripts, they lack
an existing acoustic segmentation for our model to correct. To account for this, we generate synthetic
acoustic segmentation by explicitly modeling two common error modes
of ASR acoustic segmentation: under- and over-segmentation.

\begin{figure*}[t]
\center
\noindent\fbox{%
\resizebox{0.84\textwidth}{!}{\parbox{\textwidth}{%
    \textbf{Acoustic Segmentation (Over-segmentation):}

\begin{center}    
  ARE YOU OKAY $\blacksquare$ AGENT  SCULLY $\blacksquare$ YOU KIND OF
            SOUNDED  A  $\blacksquare$  LITTLE  SPOOKY $\blacksquare$\\
   \end{center}       
   
   \textbf{Corrected Sentence Segmentation:}
   
   \begin{center}    
 ARE YOU OKAY AGENT  SCULLY $\blacksquare$ YOU KIND OF
            SOUNDED  A  LITTLE  SPOOKY $\blacksquare$ \\~\\
   \end{center}  
   
   \textbf{Acoustic Segmentation (Under-segmentation):}
   \begin{center}
        NO IS HE IN SOME KIND OF TROUBLE $\blacksquare$
   \end{center}
   
      \textbf{Corrected Sentence Segmentation:}

\begin{center}
         NO $\blacksquare$ IS HE IN SOME KIND OF TROUBLE $\blacksquare$
   \end{center}

    }}}
    \caption{Example acoustic segmentation errors and their corrections. $\blacksquare$ indicates a segment boundary. }
    \label{fig:newexample}
\end{figure*}

We evaluate the downstream MT performance in a larger SLT pipeline, and show improvements in translation quality
when using our segmentation model to correct the
acoustic segmentation provided by ASR.
 We also extrinsically evaluate our improved SLT pipeline as part of a document-level cross-lingual information retrieval (CLIR) task, where we show that improvements in ASR segmentation also lead to improved relevance of search results.
 We report results for
nine translation settings: Bulgarian (BG) to English, Lithuanian (LT) to English, and Farsi (FA) to English, and when using either phrase-based, statistical MT (SMT) or one of two neural
MT (NMT) models. We finally perform an ablation study to examine the effects of our synthetic acoustic boundaries and our over- and under-segmentation noise.

This paper makes the following contributions. \textit{(i)}~We propose the use of subtitles as a proxy dataset for correcting ASR acoustic segmentation and 
\textit{(ii)}~a method for adding synthetic acoustic utterance
segmentations to a subtitle dataset, as well as
\textit{(iii)}~a simple neural tagging model for 
correcting ASR acoustic segmentation before use in an MT pipeline. %
\textit{(iv)}~Finally, we show downstream performance increases on MT and document-level CLIR tasks, especially for more syntactically complex segments.

\section{Related Work}
Segmentation in SLT has been studied quite extensively in high-resource settings. Early work used kernel-based SVM models to predict sentence boundaries using language model probabilities along with prosodic features such as pause duration  \cite{Matusov07improvingspeech,rao2007optimizing} and part-of-speech features derived from a fixed window size \cite{Sridhar2013SegmentationSF}. Other work has modeled the problem using hidden markov models \cite{10.1016/S0167-6393(00)00028-5,Gotoh00sentenceboundary,ChristensenPunctAnnotation,Kim2001TheUO} and conditional random fields \cite{liu-etal-2005-using,lu-ng-2010-better}.

More recent segmentation work uses neural architectures, such as LSTM \cite{kit2018} and Transformer models \cite{kit2019}. These models benefit from the large training data available for high-resource languages. For example, the TED corpus \cite{cettoloEtAl:EAMT2012} for SLT from English to German includes about 340 hours of well-transcribed data.
To our knowledge, such datasets do not exist for the languages we are interested in.
\citet{wan2020subtitles} develop a segmentation model in our setting using subtitles; however, they do not take into account explicit modeling of segmentation errors and show only minimal and intermittent improvements in downstream tasks. %

Recent work has increasingly focused on end-to-end models of SLT in a high-resource setting, since these systems reduce error propagation and latency when compared to cascaded approaches \cite{Weiss2017SequencetoSequenceMC,Vila2018EndtoEndST,sperber-etal-2019-attention,gaido-etal-2020-end,bahar-etal-2020-start,lakumarapu-etal-2020-end}. In spite of these advantages, end-to-end systems have only very recently achieved competitive results due to the limited amount of parallel data for speech translation as compared to the data that is available to train ASR systems and translation systems separately \cite{gaido-etal-2020-end,ansari-etal-2020-findings}.

\section{Problem Definition}

\begin{figure*}[t]
    \centering

    \noindent\resizebox{\textwidth}{!}{\fbox{%
    \parbox{1.02\textwidth}{%
    \textbf{Reference Segmentation:}
    
    \begin{center}
        YEAH $\blacksquare$ THE HOLIDAY MARKET IS TOO BUSY $\blacksquare$ YES $\blacksquare$
        \end{center}

    \textbf{Synthetic Acoustic Segmentation Generation:} \\
    $\begin{array}{rrccccccccccccccccc}
    \toprule
    \textrm{Model Output: } & \ctags  & = & 1 & 0 & 0 & 0 & 0 & 0 & 1 & 1  \\
    \midrule
     \textrm{(Under-seg.) } & \utags & = & 1 & - & - & - & - & - & \textbf{0} & 1    \\
     \textrm{(Over-seg.) } & \otags& = & - & 0 & 0 & \textbf{1} & 0 & \textbf{1} & - & -   \\
    \midrule
  \multirow{2}{*}{\textrm{Model Input: }}    & \atags & = & 1 & 0 & 0 & 1 & 0 & 1 & 0 & 1 \\
    & \tokens & = & YEAH & THE & HOLIDAY & MARKET & IS & TOO & BUSY & YES   \\
    \bottomrule

    \end{array}
    $

    \textbf{Synthetic Acoustic Segmentation ($\atags$):}
    
    \begin{center}
        YEAH $\blacksquare$ THE HOLIDAY MARKET $\blacksquare$ IS TOO $\blacksquare$ BUSY YES $\blacksquare$
    \end{center}
    
    }%
    }}
    \caption{Example of synthetic acoustic segmentation ($\atags$) creation. For each training datapoint, we have as model input the tokens $\tokens$ and the corresponding model output sentence boundary labels $\ctags$. To generate the synthetic acoustic segmentation ($\atags$), we apply under-segmentation ($\utags$) and over-segmentation ($\otags$) noise to $\ctags$. Dashes indicate the tokens where the particular noise is not applicable. Bold indicates the changed labels due to the noise. We generate the additional input $\atags$ by combining both $\utags$ and $\otags$.
    }
    \label{fig:synthexample}
\end{figure*}

We treat the ASR acoustic segmentation problem as a sequence tagging
problem %
\cite{Stolcke1996AutomaticLS}.
Unlike a typical tagging problem, which aims to tag a single 
input sequence, our input is a pair of aligned sequences of
$\numToks$ items, $\tokens=\left[\token_1,\ldots,\token_\numToks\right]$
and $\atags = \left[\atag_1,\ldots,\atag_\numToks \right]$ where
$\tokens$ and $\atags$ are the ASR tokens and acoustic segmentation respectively.
The tokens $\token_i$ belong to a finite vocabulary $\tokVocab$, while the 
acoustic segmentation boundary tags are binary, i.e., $\atag_i \in \binspace{}$, where $\atag_i=1$
indicates that the ASR acoustic segmentation placed a boundary between 
tokens $\token_i$ and $\token_{i+1}$. The goal is to predict a corrected segment boundary tag
sequence $\ctags = \left[\ctag_1,\ldots,\ctag_\numToks \right] \in \binspace{\numToks}$
from $\tokens$ and $\atags$. 

We do this by learning a probabilistic mapping from token/segmentation sequences
to corrected segmentation $\model(\cdot|\tokens,\atags;\params) : \binspace{\numToks} \rightarrow (0,1)$ where $\model$ is a neural tagging model
with parameters $\params$. While $\atags$ are produced solely from acoustic cues,
$\model$ can take advantage of both the acoustic information (via $\atags$) as 
well as syntactic/semantic cues implicit in $\tokens$.

\section{Generating Training Data from Subtitles}

One of our primary contributions is a method for converting subtitle data into suitable
training data for an ASR segmentation correction model. The subtitle data contains speech-like utterances of dialogue between characters in film and television shows. For the purposes of this paper, we do not use information about speaker identity, only the text and information about segmentation.
We obtain the ground truth output 
label segmentation $\ctags$ by segmenting the subtitle text on
sentence final punctuation.\footnote{Set of sentence final punctuation: \{\textbf{ ( ) : - ! ? . }\}.} We remove the punctuation
but keep the implied label sequence to obtain the input token sequence $\tokens$
and ground truth output label segmentation $\ctags$. However, we  do not have acoustic segmentation available for $\tokens,\ctags$ pairs derived from subtitle data, which we will need as additional input if our model is to learn to correct acoustic segmentation provided by an ASR component. We thus create
a synthetic acoustic segmentation sequence
$\atags$
as input
by adding two types of 
noise to $\ctags$.
Specifically, we imitate two common ASR system errors,
under-segmentation noise and over-segmentation, so that at test time the model can correct those errors.

\paragraph{Under-segmentation Noise} In the ASR 
model, under-segmentation occurs  when pauses between words are brief, and the resulting ASR output is an utterance that could ideally be split into multiple sentence-like segments.
We simulate this by adding \textit{under-segmentation noise} which
converts ground truth segmentation boundaries $\ctag_i=1$, to $\ctag_i=0$ with 
probability $\usegprob$ and leaves $\ctag_i=0$ unchanged. 

\paragraph{Over-segmentation Noise} Over-segmentation occurs in an ASR 
model when a speaker takes a longer pause in the middle of what could
be interpreted as a contiguous sentence-like utterance. \textit{Over-segmentation noise} is simulated by inserting random segment boundaries within an utterance. That is, with probability
$\osegprob$ we convert a non-boundary tag
$\ctag_i=0$
to $\ctag_i=1$, while leaving all $\ctag_i=1$
unchanged.

\paragraph{Synthetic Segmentation Input Generation} We can then sample a synthetic acoustic segmentation sequence $\atags$ from the following
distribution,
\[ \atag_i \sim \begin{cases} \bernoulli{\usegprob} & \textrm{if $\ctag_i=1$}\\ \bernoulli{\osegprob} & \textrm{otherwise} \end{cases} \]
for $i \in \{1,\ldots,\numToks\}$.
This can be thought of as dropout applied to the correct label sequence $\ctags$.
See \autoref{fig:synthexample} for an example. Our proposed segmentation correction model will learn to denoise the input segmentation sequence $\atags$ and produce the corrected sequence $\ctags$.

\section{Model}

We employ a Long Short-Term Memory (LSTM)-based model architecture for this task \cite{HochSchm97}. Given an input sequence of ASR tokens $\tokens=\left[\token_1,\ldots,\token_\numToks\right]$ along with corresponding ASR segmentation sequence $\atags = \left[\atag_1,\ldots,\atag_\numToks \right]$, we first get an embedding representation $\emb_i \in \reals{316}$ for each token as follows:
\[\emb_i = G(\token_i) \oplus F( \atag_i)\]
where $G \in \reals{\vert\tokVocab\vert\times300}$ and $F \in \reals{2\times16}$ are embedding lookup tables, and $\oplus$ is the concatenation operator. We initialized $G$ with FastText embeddings pre-trained on Common Crawl data \cite{mikolov2018advances}. $F$ is randomly initialized. %

We pass the embedding sequence through a two-layer bi-directional LSTM, with $512$ hidden units each, to get the contextual representation $\hid_i \in \reals{1024}$ for each token as follows:
\[\hid_i = \flstm(\emb_i) \oplus \blstm(\emb_i)\]
where $\flstm$ and $\blstm$ are the forward direction and backward direction LSTMs respectively.

Each output state $\hid_i$ is then passed through a linear projection layer with a logistic sigmoid to compute the probability of a segment boundary 
$p(\ctag_i=1|\hid_i;\params)$. The log-likelihood of a corrected segmentation boundary sequence is $\log p(\ctags|\tokens,\atags;\params) = \sum_{i=1}^n \log p(\ctag_i|\hid_i;\params)$. We fit the parameters, $\params$, 
by approximately minimizing the negative log-likelihood 
on the training set $\corpus$, $\mathcal{L}(\params) =  - \frac{1}{\lvert \corpus\rvert}\sum_{\left(\tokens,\atags,\ctags\right) \in \corpus  }\log \model\left(\ctags|\atags,\tokens;\params \right)$,
using mini-batch stochastic gradient descent.

\newcommand{\BUILD}{BUILD}
\newcommand{\ANALYSIS}{ANALYSIS}
\newcommand{\DEV}{DEV}
\newcommand{\EVAL}{EVAL}

\section{Datasets}

\begin{table*}
\centering
\begin{tabular}{c ccc ccc  ccc ccc  ccc }
\toprule
 \multirow{2}{*}{Lang.} & \multicolumn{3}{c}{OpenSubtitles} & \multicolumn{3}{c}{\BUILD~Train} & \multicolumn{3}{c}{\BUILD~Valid}  \\
 \cmidrule(lr){2-4}\cmidrule(lr){5-7}\cmidrule(lr){8-10}
 & D & I & S & D & I & S & D & I & S  \\
 \midrule
BG & 2000 & \phantom{0,}459,301 & \phantom{0}3.95 & 352 & \phantom{0}7,723 & 3.22 & 108 & 2,472 & 3.14    \\
FA & 2000 & \phantom{0,}120,039 & \phantom{0}3.88 & 302 & \phantom{0}6,707 & 3.52 & 120 & 2,679 & 3.53  \\
LT & 1977 & \phantom{0,}165,751 & 12.99 & 484 & 11,782 & 3.40 & 112 & 2,893 & 3.31  \\
\bottomrule
\end{tabular}
\caption{Segmentation model training dataset statistics. We report number of documents (D), number of instances (I) in each dataset and average number of segments
per instance (S).}
\label{tab:dataset}
\end{table*}

\begin{table}
\centering
\begin{tabular}{c cccc  }
\toprule
 Lang. & \multicolumn{2}{c}{Test (Small)} & \multicolumn{2}{c}{Test (Large)} \\
 \cmidrule(lr){2-3} \cmidrule(lr){4-5}
 &Q & D & Q &D\\
 \midrule
BG & 300  & 634 & -- & --\\
FA & 221  & 528 & -- & -- \\
LT & 300  & 496 & 1,000   & 3,297 \\
\bottomrule
\end{tabular}
\caption{Number of queries (Q) and documents (D) in the speech retrieval test collections. } %

\label{tab:testdataset}
\end{table}

\subsection{Subtitles Dataset}
We obtain monolingual subtitle data from the
OpenSubtitles 2018 corpus \cite{lison2016opensubtitles2016}.
OpenSubtitles contains monolingual subtitles for 62 languages drawn 
from movies and television. The number of subtitle documents 
varies considerably from language to language. LT has only 1,976 documents, while BG and FA have 107,923 and 12,185 %
respectively. %
We randomly down-sample from the larger collection to 2,000 documents to ensure our segmentation correction models are all trained with similar amounts of data.%

Treating the subtitles for a complete television
episode or movie as the source of a single training instance $(\tokens,\atags,\ctags)$
introduces some complications because they are usually quite long relative to typical
SLT system input. To better match our evaluation conditions, we arbitrarily split each document into $M$ instances, where the length $l$ in tokens for each instance $m$ is sampled from $L \sim U(1,100)$, i.e. uniformly from 1 to 100 tokens. This range was determined to
to approximate the length distribution of our evaluation datasets.

See \autoref{tab:dataset} for statistics on the number of training instances created as well as the average number of sentence segments per instance. Note that even though the number of subtitle documents is close to equal, the documents can vary considerably in length. As result, the BG dataset has more than twice the training instances of FA or LT.
In some cases, an instance may contain only a few words that do not constitute a sentence, and such instances would have no segment boundaries; this helps
prevent the model from learning pathological solutions such as always inserting a 
segment boundary at the end of the sequence.

Since we do not evaluate the segmentation directly on OpenSubtitles,
we split the available data into training and development partitions, with 90\% of the instances
in the training set.

\subsection{Speech Retrieval Dataset}

For extrinsic evaluation of ASR segments, we use the speech retrieval dataset
from the MATERIAL\footnote{\url{www.iarpa.gov/index.php/research-programs/material}} program. The goal of MATERIAL is to develop systems that can retrieve text and speech documents in low-resource languages that are relevant to a given query in English. 
To bootstrap speech retrieval systems in low-resource languages, MATERIAL
collects 
BG, FA, and LT speech training data for ASR systems, as well as additional separate collections of BG, FA, and LT speech documents along with their relevance judgments for a set of English language queries.
Since the retrieval of speech documents requires a cascade of ASR, MT, and CLIR systems, the MATERIAL data allows us to measure the impact of ASR segmentation on both the translation quality, as well as the downstream retrieval system.
The data partitions in MATERIAL are numerous and to avoid confusion, we briefly describe them here.

The \textit{\BUILD}~partition contains a small amount of ASR training 
and development data for BG, FA, and LT, i.e. audio files paired with reference transcripts. 
We use the \BUILD~data for fine-tuning our subtitle trained model. We apply the same synthetic acoustic
segmentation generation procedure to this collection as we do to the subtitle data when using it for fine-tuning. See 
\autoref{tab:dataset} for dataset statistics.

The \textit{Test (Small)} partition contains audio documents and a set of 
English language queries and relevance judgements for those queries.
At test time,
we use the acoustic segmentation provided by the ASR system as the input $\atags$ instead of generating acoustic label sequences.
Additionally, roughly half of the audio documents in this collection include ground-truth transcriptions
and translations to English, which allows us to evaluate MT.

The \textit{Test (Large)} partition is similar to the Test (Small) partition, but much  
bigger in size. There are no transcripts or translations, so it can be used only to evaluate CLIR. The Test (Large)~partition is available only for LT.

We use the translated portion of Test (Small) as a test set for MT and both Test (Small) and Test (Large) as extrinsic test sets for CLIR. 
The statistics of the MATERIAL partitions
can be found in \autoref{tab:testdataset}.\footnote{The official MATERIAL collections are named ANALYSIS+DEV and EVAL, but we refer to them as Test (Small) and Test (Large) to avoid confusion.}

The speech retrieval datasets come from three domains: news broadcast, topical broadcast such as podcasts, and conversational speech from multiple low-resource languages. 
Some speech documents have two speakers, with 
each speaker on a separate channel, i.e., %
completely isolated from the other speaker. When
performing segmentation we treat each channel independently, creating a separate (re-segmented) ASR output for each channel.
To create the document transcript for
MT, we merge the two output sequences by sorting 
the token segments based on their wall-clock start
time.

\section{Experiments}

\subsection{Segmentation Model Training}

For all datasets, we tokenize all data with Moses \cite{koehn-etal-2007-moses}. To improve performance on out of vocabulary words, we use Byte-Pair-Encoding \cite{sennrich-etal-2016-neural} with 32,000 merge operations to create subwords for each language.

We then train the segmentation model on the subtitle dataset.
When creating $\atag$ sequences on the subtitles data,
we set under- and over-segmentation noise to $\usegprob=0.25$ and $\osegprob=0.25$ respectively.\footnote{Values for $\usegprob$ and $\osegprob$ were determined by grid-search over $\{0.25, 0.5, 0.75 \}$ that minimized loss on the BUILD Valid data.} 
We use the Adam optimizer \citep{kingma14} with learning rate of 0.001. We use early stopping on the validation loss of the OpenSubtitles validation set to select the best stopping
epoch for the segmentation model. %

We further fine-tune this model on the BUILD partition to expose the model to some in-domain training data. The data is similarly prepared as OpenSubtitles. We use early stopping on the development loss of this partition. %

\subsection{ASR-Segmentation-MT-CLIR Pipeline}

We evaluate our segmentation correction model in the context of a CLIR pipeline for 
retrieving audio documents in
BG, FA, or LT that are relevant to English queries. We refer to the three languages BG, FA, and LT as \textit{source languages}. This pipeline uses ASR to convert source language audio documents to source language text transcripts, and MT to translate the source language transcripts into English transcripts. Then a monolingual English IR component is used to return source language documents that are relevant to the issued English queries.
We insert
our segmentation correction model into this pipeline between the ASR and MT components, i.e. \textit{(i)} ASR to \textit{(ii)} Segmentation Correction, to  \textit{(iii)} MT to 
\textit{(iv)} IR.
For clarity we reiterate, the segmentation model takes as input a source language transcript and returns the source language transcript with corrected segmentation.

To implement the ASR, MT, and IR components, we use implementations developed by MATERIAL program participants \cite{5day}. %

\subsubsection{ASR System}

We use the ASR systems developed jointly by the University of Cambridge and the University of Edinburgh \cite{ragni18inthewild,carmantini19semisup}. 
The ASR system uses a neural network based acoustic model, trained in a semi-supervised manner on web-scraped audio data, to overcome
the small amount of training data in the \BUILD~data. Separate models 
are trained for narrow-band audio (i.e., conversational speech)
and wide-band audio (i.e. news and topical broadcast).

\subsubsection{Segmentation Correction}
At test time, given a speech document, the ASR system produces a series of acoustically derived utterances,
i.e. $\tokens^{(1)},\ldots, \tokens^{(m)}$, from this input. In our setting, the corresponding acoustic label sequence $\atags^{(i)}$ for each utterance would be zero
everywhere except the final position, i.e. $\atags^{(i)} = \left[ 0, 0, \ldots, 0, 1\right]$. If we were to process each utterance, $(\tokens^{(i)},\atags^{(i)})$, 
individually, the model may not have enough context to correct under-segmentation 
at the ends of the utterance. For example, when correcting the final token position, which by definition will precede a long audio pause, the model will only see the left-hand side of the context. To avoid this, we run our segmentation correction model 
on consecutive pairs of ASR output utterances, i.e. $\left(\tokens^{(i)} \oplus \tokens^{(i+1)} , \atags^{(i)}\oplus \atags^{(i+1)} \right)$. Under this formulation each ASR output utterance is corrected twice (except for the first and last utterances which are only corrected once), therefore we have two predictions $\hat{\ctag}^{(i,L)}_j$ and  $\hat{\ctag}^{(i,R)}_j$ for the $j$-th segment boundary. We resolve these with the logical-OR operation to obtain the final segmentation correction, i.e.  $\hat{\ctag}^{(i)}_j = \hat{\ctag}^{(i,L)}_j \vee \hat{\ctag}^{(i,R)}_j$.

Based on the segmentation corrections produced by our model, 
we re-segment the ASR output tokens and hand the resulting segments off to the MT
component where they are individually translated.

\subsubsection{MT Systems}

We evaluate with three different MT systems. We use the neural MT model developed by the University of Edinburgh  (EDI-NMT) and
the neural and phrase-based statistical MT systems from the University of 
Maryland (UMD-NMT and UMD-SMT, respectively). The EDI-NMT and UMD-NMT systems are Transformer-based models \cite{NIPS2017_7181} trained using the Marian Toolkit \cite{mariannmt} and Sockeye \cite{Sockeye:17}, respectively. UMD-NMT trains a single model for both directions of a language pair \cite{niu-etal-2018-bi}, while EDI-NMT has a separate model for each direction. UMD-SMT is trained using the Moses SMT Toolkit \cite{koehn-etal-2003-statistical}, where the weights were optimized using  MERT  \cite{och-2003-minimum}.

\subsubsection{IR System}
For the IR system, we use the bag-of-words language model
implemented in Indri \cite{strohman2005indri}. Documents and queries are both tokenized and normalized on the character level to avoid potential mismatch in the vocabulary. The queries are relatively short, typically consisting of only a few words, and they define two types of relevancy -- the conceptual queries require the relevant documents to be topically relevant to the query, while the simple queries require the relevant document to contain the translation of the query. However, no specific processing is used for these two relevance types in our experiments.

\subsection{MT Evaluation}
Our first extrinsic evaluation measures the BLEU \cite{papineni2002bleu} score of 
the MT output on the Test (Small)~sets after running our segmentation correction model, where we have ground truth 
reference English translations. We refer to our model trained only on the BUILD data as \textit{Seg}, and our subtitle-trained model as \textit{Seg + Sub}. As our baseline, we compare the same pipeline using the segmentation produced by the acoustic model of the ASR system,
denoted \textit{Acous}.

Since each segmentation model  produces segments with
different boundaries, we are unable to use
BLEU directly to compare to the reference sen-
tences. Therefore, we concatenate all segments of a
document and treat them as one segment, which we refer to as ``document-level'' BLEU score.
We use SacreBLEU\footnote{\url{https://github.com/mjpost/sacrebleu}} \cite{post-2018-call} with the lowercase option due to the different casing for the reference English translation and MT output.

We also provide BLEU scores for the MT output using the reference transcriptions (Ref) to show the maximum score the system can achieve when there is no ASR or segmentation error. This represents the theoretical upper bound for our pipeline with a  perfect ASR system.

Segmentation errors (i.e., the acoustic model incorrectly segmented an utterance) and word errors (i.e., the ASR system produces an incorrect word) can both affect the downstream MT performance.
To isolate the segmentation errors from word errors, we align the ASR output tokens to the reference transcriptions by timecode in order to obtain a reference system that has no segmentation errors, but does have
transcription errors. This represents a more
realistic ceiling for our model because while we can correct segmentation, we cannot correct word errors. We refer to this system in the results section as Align.

\subsection{Document-Level CLIR Evaluation}

Our second extrinsic evaluation is done on the MATERIAL CLIR task.
We are given English queries and asked to retrieve audio documents in 
either BG, FA, or LT. In our setup, we only search over the English translations of the segmented transcripts 
produced by our pipeline, i.e., we do not translate the English query into the other languages or search the audio signal directly. 
We evaluate the performance of CLIR using the Maximum Query Weighted Value (MQWV) from the ground-truth query-relevance judgements for documents in the Test (Small \& Large)~collections.
MQWV, which is a variant of the official MATERIAL program metric %
called Actual Query Weighted Value~\cite[AQWV]{aqwv}, is a recall-oriented rank metric that measures
how well we order the retrieval collection with respect to query relevance.

AQWV is calculated as the average of $1 - (P_{m} + \beta * P_{fa})$ for each query, where $P_{m}$ is the probability of misses, $P_{fa}$ is the probability of false alarms, and $\beta$ is a hyperparameter. The maximum possible value is 1 and the minimum value is given by $- \beta$.
In our experiments $\beta$ it is set to 40.
AQWV
thus not only depends on the ranking of the documents but also on $\beta$.

Additionally, AQWV is sensitive to the threshold used by the IR system to determine document relevance. %
To avoid the tuning of thresholds, we report MQWV which %
is calculated for the optimal threshold; in our experiments this threshold is estimated over the ranks of the documents. Thus, MQWV doesn't depend on the ability to estimate the threshold and only depends on the quality of the document ranking for a given query.

\begin{table}[t]
\centering
\begin{tabular}{ llccc}
\toprule
\multirow{2}{*}{Lang.} & \multirow{2}{*}{Model} & EDI & UMD  & UMD \\
& & NMT & NMT & SMT\\
\midrule
\multirow{4}{*}{BG} & Acous. & 20.48\phantom{*} & 20.39\phantom{*} & \textbf{21.24} \\
& Seg & 22.38* & 23.35* & 21.23 \\
& Seg + Sub &  \textbf{24.73*} & \textbf{25.92*} & 21.23 \\
\cmidrule(lr){2-5}
& Align & 24.98\phantom{*} & 27.81\phantom{*} & 21.29 \\
& Ref & 43.75\phantom{*} & 35.40\phantom{*} & 29.50 \\
\midrule
\multirow{3}{*}{FA} & Acous. & 5.35\phantom{*} & 6.26\phantom{*} & \textbf{4.54} \\
& Seg & 5.32\phantom{*} & 6.22\phantom{*} & 3.28 \\ 
& Seg + Sub &  \textbf{6.47*} & \textbf{6.83*} & 4.50 \\
\cmidrule(lr){2-5}
& Align &  7.67\phantom{*} & 7.02\phantom{*} & 4.59 \\
& Ref & 17.08\phantom{*} & 11.24\phantom{*} & 7.76 \\
\midrule
\multirow{4}{*}{LT} & Acous. & 15.20\phantom{*} & \textbf{8.38}\phantom{*} & 14.76 \\
& Seg & 15.18\phantom{*} & 8.34\phantom{*} & 14.76 \\
& Seg + Sub & \textbf{15.22}\phantom{*} & 8.33\phantom{*} & 14.76 \\
\cmidrule(lr){2-5}
& Align & 15.60\phantom{*} & 8.71\phantom{*} & 14.84 \\
& Ref &  20.40\phantom{*} & 11.94\phantom{*} & 21.30  \\
\bottomrule
\end{tabular}
\caption {Document-level BLEU scores on ANALYSIS set. * represents statistical significance when compared to Acous. at the 0.05 level.
We show the result of translations using the original acoustic segmentation (Acous.), our model trained only on the BUILD dataset (Seg), and our full model (Seg + Sub). For reference, we provide the scores of translation on ASR tokens aligned to the reference transcription segmentation (Align), and the reference transcription (Ref). }
\label{bleu}
\end{table}

\section{Results}
\subsection{MT}
\autoref{bleu} shows the results of the MT evaluation. The best non-reference system for each language and MT system is in bold. We compute statistical significance against the 
acoustic (\textit{Acous.}) segmentation baseline using Welch's T Test \cite{10.1093/biomet/34.1-2.28}.
Our subtitle-based segmentation model (\textit{Seg + Sub}) consistently improves BLEU scores of NMT models for BG and FA, while not making significant differences in SMT.
This echoes prior work \cite{khayrallah-koehn-2018-impact,rosales-nunez-etal-2019-comparison}
suggesting SMT models are more robust to noisy inputs than neural models.

In 6 out of 9 cases, we see that adding the subtitles data improves over using only the \BUILD~data. Of the remaining cases, the scores remain similar (i.e., it doesn't hurt the model).
Training on the \BUILD~data alone improves BG NMT models, but for SMT and the other languages, it either makes no difference or is worse than the acoustic model.

Comparing Seg + Sub with Align in all languages, we see that there is only a small gap between the two. This suggests that our model is nearing the ceiling on what correcting segmentation can do to improve downstream MT.
Furthermore, on LT where our model offers only small or no improvement, we see that the original acoustic segmentation is almost performing as well as Align. This suggests that there is relatively little room for improving LT MT by correcting sentence boundaries alone.

\subsection{Document-Level CLIR}
MQWV on the Test (Small) and Test (Large) partitions are shown in \autoref{mqwvopus} and \autoref{aqwveval} respectively. On the Test (Small) partition, we see that our segmentation model improves the CLIR performance over the acoustic segmentation in 7 out of 9 cases. On the Test (Large) partition, we see that our segmentation model improves downstream retrieval performance consistently across all three MT systems. We note that while we measure the downstream retrieval performance separately for each MT system, a real-world CLIR system could perform IR over the union of multiple MT systems, which could yield even
further improvements in retrieval performance \cite{zhang-etal-2020-2019}.

\begin{table}[t]
\centering
\begin{tabular}{ ll ccc}
\toprule
 \multirow{2}{*}{Lang.} & \multirow{2}{*}{Model} & EDI & UMD & UMD \\
  & & NMT & NMT & SMT \\
\midrule
\multirow{3}{*}{BG} & Acous. & 0.173 & 0.134 & 0.164  \\
& Seg + Sub &  \textbf{0.177} & \textbf{0.180} & 0.164 \\
\midrule
\multirow{3}{*}{FA} & Acous. & 0.040 & 0.039 &  0.046 \\
& Seg + Sub & \textbf{0.071} & \textbf{0.042} & \textbf{0.145} \\
\midrule
\multirow{3}{*}{LT} & Acous. & 0.128 & \textbf{0.067} & 0.157 \\
& Seg + Sub &  \textbf{0.136} & 0.060 & \textbf{0.172} \\
\bottomrule
\end{tabular}
\caption {MQWV scores on Test (Small) set.}
\label{mqwvopus}
\end{table}

\begin{table}[ht]
\label{mqwvltopus}
\centering
\begin{tabular}{ llccc}
\toprule
 \multirow{2}{*}{Lang.} & \multirow{2}{*}{Model} & EDI & UMD & UMD \\
     & & NMT & NTM & SMT \\
\midrule
\multirow{2}{*}{LT} & Acous. & 0.292 & 0.175 & 0.328 \\
& Seg + Sub & \textbf{0.293} & \textbf{0.179} & \textbf{0.399} \\
\bottomrule
\end{tabular}
\caption {MQWV scores on the Test (Large) set.}
\label{aqwveval}
\end{table}

\subsection{Complexity Analysis} \label{sec:complexity}

\begin{table}[t]
\centering
\begin{tabular}{ ll ccc}
\toprule
 ARI & \multirow{2}{*}{Model} & EDI & UMD \\
  Quartile & & NMT & NMT \\
\midrule
\multirow{2}{*}{Q1} & Acous. & 17.78 & \textbf{24.70} \\
& Seg + Sub & \textbf{17.92} & 11.62 \\
\midrule
\multirow{2}{*}{Q2} & Acous.  & 20.76 & 22.09 \\
& Seg + Sub & \textbf{26.89} & \textbf{31.24} \\
\midrule
\multirow{2}{*}{Q3} & Acous. & 22.87 & 20.38 \\
& Seg + Sub & \textbf{29.96} & \textbf{33.22} \\
\midrule
\multirow{2}{*}{Q4} & Acous.  & 23.41 & 21.00 \\
& Seg + Sub & \textbf{29.87} & \textbf{35.24} \\
\bottomrule
\end{tabular}
\caption {Bulgarian BLEU scores on Test (Small) (transcribed portion) when separated into quartiles by sentence complexity (as measured by ARI).}
\label{tab:bleu_quartiles}
\end{table}

We hypothesize that the effects of improved segmentation should be more pronounced for more complex utterances with more opportunities to misplace boundaries. Therefore, we calculate a measure of sentence complexity, the Automated Readability Index (ARI) \citep{ari}, for all documents in Test (Small)\footnote{Only the transcribed portion with reference translations. For each MT system, we compute ARI on the document translation using the reference transcription.} and examine the performance of our \textit{Sub} model on MT. We separate the documents into quartiles based on their calculated ARI, where a higher ARI (and thus a higher quartile) indicates a more complex document, and present the average document-level BLEU score for each quartile in \autoref{tab:bleu_quartiles}. In the interest of space, we present results for Bulgarian and for NMT, and defer other languages and SMT to \autoref{complexity}. We see that the most dramatic gains in BLEU occur for documents in the third and fourth quartiles, which matches our intuition. In other words, our segmentation model most improves the translation quality of more syntactically complex segments. 

\section{Ablation Study}
We perform an ablation study on two components  in our proposed model, \textit{(i)} the use of acoustic segmentation boundary labels $\atags$ as input  and \textit{(ii)} training with a combination of over- and under-segmentation noise. 
 We use the same training and evaluation process and only modify the affected component. We perform our ablation on the BG MT task, since it had a wider range of improvements than the other languages.

\paragraph{Use of Acoustic Segmentation Boundaries.}
We train a segmentation model using only ASR output tokens $\tokens$ as input
without the the ASR segmentation sequence $\atags$. For this model, we modify the embedding representation $\mathbf{e}_i$ so that we do not use $F$:
\[\emb_i = G(\token_i)\]
This model, which we refer to as \textit{Lex.}, must exclusively use the lexical information
of the ASR token sequence $\tokens$ to make predictions.

\paragraph{Over-segmentation and Under-segmentation.}
The two segmentation problems of the system may have different impact on the MT system. To see their
individual effects, we train two models where the synthetic acoustic segmentation boundary sequence $\atags$ is created using only under-segmentation
or over-segmentation noise. We refer to those models
as \textit{Lex. + Under} and \textit{Lex. + Over} respectively.

\begin{table}[t]
\centering
\begin{tabular}{ lccc}
\toprule
\multirow{2}{*}{Model} & EDI & UMD & UMD \\
  & NMT & NMT & SMT \\
\midrule
Acous. & 20.48 & 20.39 & 21.24\\
Lex. & 23.96 & 24.97 & \textbf{21.58} \\
Lex. + Under & 24.27 & 25.84 & 21.48 \\
Lex. + Over & 24.69 & 25.88 & 21.23 \\
Full &  \textbf{24.73} & \textbf{25.92} & 21.23 \\

\bottomrule
\end{tabular}
\caption {Document-level BLEU score for the models in ablation studies. We provide the acoustic model (Acous.) and our proposed model (Full). }
\label{ablbleu}
\end{table}

\paragraph{Results}

\autoref{ablbleu} shows the effects of the model ablations on MT system BLEU score. On both NMT systems, we see that there is a roughly 1 point improvement on BLEU when including the ASR segmentation boundaries
as input. For both NMT models we also find that over-segmentation noise helps slightly more than adding under-segmentation noise, but that these additions
are complementary, i.e. the full model does best
overall.
For SMT, we 
surprisingly
find that model without acoustic segmentation boundary input does best. The overall 
difference between the acoustic (\textit{Acous.}) baseline and any of 
the segmentation correction models is small compared 
to the gains had on NMT. This again suggests that SMT is more robust to changes in segmentation.

\section{Conclusion}

We propose an ASR segmentation correction model for improving SLT pipelines. Our model makes use of subtitles data as well 
as a simple model of acoustic segmentation error to train
an improved ASR segmentation model. 
We demonstrate downstream improvements on MT and CLIR tasks.
In future work, we would like to find a better segmentation error model that works well in conjunction with SMT systems in addition to NMT systems.

\section*{Acknowledgements}
This research is based upon work supported inpart by the Office of the Director of National Intelligence (ODNI), Intelligence Advanced Research Projects Activity (IARPA), via contract \#FA8650-17-C-9117. The views and conclusions contained herein are those of the authors and should not be interpreted as necessarily representing the official policies, either expressed or implied, of ODNI, IARPA, or the U.S. Government. The U.S. Government is authorized to reproduce and distribute reprints for governmental purposes not withstanding any copyright annotation therein.

\bibliography{segmentation}

\begin{thebibliography}{46}
\expandafter\ifx\csname natexlab\endcsname\relax\def\natexlab#1{#1}\fi

\bibitem[{Ansari et~al.(2020)Ansari, Axelrod, Bach, Bojar, Cattoni, Dalvi,
  Durrani, Federico, Federmann, Gu, Huang, Knight, Ma, Nagesh, Negri, Niehues,
  Pino, Salesky, Shi, St{\"u}ker, Turchi, Waibel, and
  Wang}]{ansari-etal-2020-findings}
Ebrahim Ansari, Amittai Axelrod, Nguyen Bach, Ond{\v{r}}ej Bojar, Roldano
  Cattoni, Fahim Dalvi, Nadir Durrani, Marcello Federico, Christian Federmann,
  Jiatao Gu, Fei Huang, Kevin Knight, Xutai Ma, Ajay Nagesh, Matteo Negri, Jan
  Niehues, Juan Pino, Elizabeth Salesky, Xing Shi, Sebastian St{\"u}ker, Marco
  Turchi, Alexander Waibel, and Changhan Wang. 2020.
\newblock \href {https://doi.org/10.18653/v1/2020.iwslt-1.1} {{FINDINGS} {OF}
  {THE} {IWSLT} 2020 {EVALUATION} {CAMPAIGN}}.
\newblock In \emph{Proceedings of the 17th International Conference on Spoken
  Language Translation}, pages 1--34, Online. Association for Computational
  Linguistics.

\bibitem[{Bahar et~al.(2020)Bahar, Wilken, Alkhouli, Guta, Golik, Matusov, and
  Herold}]{bahar-etal-2020-start}
Parnia Bahar, Patrick Wilken, Tamer Alkhouli, Andreas Guta, Pavel Golik, Evgeny
  Matusov, and Christian Herold. 2020.
\newblock \href {https://doi.org/10.18653/v1/2020.iwslt-1.3}
  {Start-{B}efore-{E}nd and {E}nd-to-{E}nd: {N}eural {S}peech {T}ranslation by
  {A}pp{T}ek and {RWTH} {A}achen {U}niversity}.
\newblock In \emph{Proceedings of the 17th International Conference on Spoken
  Language Translation}, pages 44--54, Online. Association for Computational
  Linguistics.

\bibitem[{Carmantini et~al.(2019)Carmantini, Bell, and
  Renals}]{carmantini19semisup}
Andrea Carmantini, Peter Bell, and Steve Renals. 2019.
\newblock \href {https://doi.org/10.21437/Interspeech.2019-2623}
  {{Untranscribed Web Audio for Low Resource Speech Recognition}}.
\newblock In \emph{Proceedings of Interspeech 2019}, pages 226--230.

\bibitem[{Cettolo et~al.(2012)Cettolo, Girardi, and
  Federico}]{cettoloEtAl:EAMT2012}
Mauro Cettolo, Christian Girardi, and Marcello Federico. 2012.
\newblock \href {https://www.aclweb.org/anthology/2012.eamt-1.60} {{WIT}$^3$:
  {W}eb {I}nventory of {T}ranscribed and {T}ranslated {T}alks}.
\newblock In \emph{Proceedings of the 16th Annual conference of the European
  Association for Machine Translation}, pages 261--268, Trento, Italy. European
  Association for Machine Translation.

\bibitem[{Cho et~al.(2017)Cho, Niehues, and Waibel}]{segmentationimportance}
Eunah Cho, Jan Niehues, and Alex Waibel. 2017.
\newblock \href {https://doi.org/10.21437/Interspeech.2017-1320} {{NMT}-{B}ased
  {S}egmentation and {P}unctuation {I}nsertion for {R}eal-{T}ime {S}poken
  {L}anguage {T}ranslation}.
\newblock In \emph{Proceedings of Interspeech 2017}, pages 2645--2649.

\bibitem[{Christensen et~al.(2001)Christensen, Gotoh, and
  Renals}]{ChristensenPunctAnnotation}
Heidi Christensen, Yoshihiko Gotoh, and Steve Renals. 2001.
\newblock \href
  {https://www.isca-speech.org/archive_open/archive_papers/prosody_2001/prsr_006.pdf}
  {Punctuation {A}nnotation using {S}tatistical {P}rosody {M}odels}.
\newblock In \emph{Proceedings of ISCA Workshop on Prosody in Speech
  Recognition and Understanding}, pages 35--40.

\bibitem[{{Cross Vila} et~al.(2018){Cross Vila}, Escolano, Fonollosa, and {R.
  Costa-Jussà}}]{Vila2018EndtoEndST}
Laura {Cross Vila}, Carlos Escolano, José A.~R. Fonollosa, and Marta {R.
  Costa-Jussà}. 2018.
\newblock \href {https://doi.org/10.21437/IberSPEECH.2018-13} {{End-to-End
  Speech Translation with the Transformer}}.
\newblock In \emph{Proceedings of IberSPEECH 2018}, pages 60--63.

\bibitem[{Gaido et~al.(2020)Gaido, Di~Gangi, Negri, and
  Turchi}]{gaido-etal-2020-end}
Marco Gaido, Mattia~A. Di~Gangi, Matteo Negri, and Marco Turchi. 2020.
\newblock \href {https://doi.org/10.18653/v1/2020.iwslt-1.8} {End-to-end
  {S}peech-{T}ranslation with {K}nowledge {D}istillation: {FBK}@{IWSLT}2020}.
\newblock In \emph{Proceedings of the 17th International Conference on Spoken
  Language Translation}, pages 80--88, Online. Association for Computational
  Linguistics.

\bibitem[{Gotoh and Renals(2000)}]{Gotoh00sentenceboundary}
Yoshihiko Gotoh and Steve Renals. 2000.
\newblock \href
  {https://isca-speech.org/archive_open/archive_papers/asr2000/asr0_228.pdf}
  {Sentence {B}oundary {D}etection in {B}roadcast {S}peech {T}ranscripts}.
\newblock In \emph{Proceedings of ISCA Workshop: Automatic Speech Recognition:
  Challenges for the new Millennium ASR-2000}, pages 228--235.

\bibitem[{Hieber et~al.(2018)Hieber, Domhan, Denkowski, Vilar, Sokolov,
  Clifton, and Post}]{Sockeye:17}
Felix Hieber, Tobias Domhan, Michael Denkowski, David Vilar, Artem Sokolov, Ann
  Clifton, and Matt Post. 2018.
\newblock \href {https://www.aclweb.org/anthology/W18-1820} {The {S}ockeye
  {N}eural {M}achine {T}ranslation {T}oolkit at {AMTA} 2018}.
\newblock In \emph{Proceedings of the 13th Conference of the Association for
  Machine Translation in the {A}mericas (Volume 1: Research Papers)}, pages
  200--207, Boston, MA. Association for Machine Translation in the Americas.

\bibitem[{Hochreiter and Schmidhuber(1997)}]{HochSchm97}
Sepp Hochreiter and Jürgen Schmidhuber. 1997.
\newblock \href {https://doi.org/10.1162/neco.1997.9.8.1735} {Long
  {S}hort-{T}erm {M}emory}.
\newblock \emph{Neural Computation}, 9(8):1735--1780.

\bibitem[{Junczys-Dowmunt et~al.(2018)Junczys-Dowmunt, Grundkiewicz, Dwojak,
  Hoang, Heafield, Neckermann, Seide, Germann, Fikri~Aji, Bogoychev, Martins,
  and Birch}]{mariannmt}
Marcin Junczys-Dowmunt, Roman Grundkiewicz, Tomasz Dwojak, Hieu Hoang, Kenneth
  Heafield, Tom Neckermann, Frank Seide, Ulrich Germann, Alham Fikri~Aji,
  Nikolay Bogoychev, Andr\'{e} F.~T. Martins, and Alexandra Birch. 2018.
\newblock \href {https://arxiv.org/abs/1804.00344} {Marian: {F}ast {N}eural
  {M}achine {T}ranslation in {C++}}.
\newblock In \emph{Proceedings of ACL 2018, System Demonstrations}, Melbourne,
  Australia.

\bibitem[{Khayrallah and Koehn(2018)}]{khayrallah-koehn-2018-impact}
Huda Khayrallah and Philipp Koehn. 2018.
\newblock \href {https://doi.org/10.18653/v1/W18-2709} {On the {I}mpact of
  {V}arious {T}ypes of {N}oise on {N}eural {M}achine {T}ranslation}.
\newblock In \emph{Proceedings of the 2nd Workshop on Neural Machine
  Translation and Generation}, pages 74--83, Melbourne, Australia. Association
  for Computational Linguistics.

\bibitem[{Kim and Woodland(2001)}]{Kim2001TheUO}
Ji{-}Hwan Kim and Philip~C. Woodland. 2001.
\newblock \href
  {http://www.isca-speech.org/archive/eurospeech\_2001/e01\_2757.html} {The use
  of {P}rosody in a {C}ombined {S}ystem for {P}unctuation {G}eneration and
  {S}peech {R}ecognition}.
\newblock In \emph{{EUROSPEECH} 2001 Scandinavia, 7th European Conference on
  Speech Communication and Technology, 2nd {INTERSPEECH} Event, Aalborg,
  Denmark, September 3-7, 2001}, pages 2757--2760. {ISCA}.

\bibitem[{Kingma and Ba(2015)}]{kingma14}
Diederik~P. Kingma and Jimmy Ba. 2015.
\newblock \href {http://arxiv.org/abs/1412.6980} {Adam: {A} method for
  {S}tochastic {O}ptimization}.
\newblock In \emph{3rd International Conference on Learning Representations,
  {ICLR} 2015, San Diego, CA, USA, May 7-9, 2015, Conference Track
  Proceedings}.

\bibitem[{Koehn et~al.(2007)Koehn, Hoang, Birch, Callison-Burch, Federico,
  Bertoldi, Cowan, Shen, Moran, Zens, Dyer, Bojar, Constantin, and
  Herbst}]{koehn-etal-2007-moses}
Philipp Koehn, Hieu Hoang, Alexandra Birch, Chris Callison-Burch, Marcello
  Federico, Nicola Bertoldi, Brooke Cowan, Wade Shen, Christine Moran, Richard
  Zens, Chris Dyer, Ond{\v{r}}ej Bojar, Alexandra Constantin, and Evan Herbst.
  2007.
\newblock \href {https://www.aclweb.org/anthology/P07-2045} {{M}oses: {O}pen
  {S}ource {T}oolkit for {S}tatistical {M}achine {T}ranslation}.
\newblock In \emph{Proceedings of the 45th Annual Meeting of the Association
  for Computational Linguistics Companion Volume Proceedings of the Demo and
  Poster Sessions}, pages 177--180, Prague, Czech Republic. Association for
  Computational Linguistics.

\bibitem[{Koehn et~al.(2003)Koehn, Och, and
  Marcu}]{koehn-etal-2003-statistical}
Philipp Koehn, Franz~J. Och, and Daniel Marcu. 2003.
\newblock \href {https://www.aclweb.org/anthology/N03-1017} {Statistical
  {P}hrase-{B}ased {T}ranslation}.
\newblock In \emph{Proceedings of the 2003 Human Language Technology Conference
  of the North {A}merican Chapter of the Association for Computational
  Linguistics}, pages 127--133.

\bibitem[{Lakumarapu et~al.(2020)Lakumarapu, Lee, Indurthi, Han, Zaidi, and
  Kim}]{lakumarapu-etal-2020-end}
Nikhil~Kumar Lakumarapu, Beomseok Lee, Sathish~Reddy Indurthi, Hou~Jeung Han,
  Mohd~Abbas Zaidi, and Sangha Kim. 2020.
\newblock \href {https://doi.org/10.18653/v1/2020.iwslt-1.7} {End-to-{E}nd
  {O}ffline {S}peech {T}ranslation {S}ystem for {IWSLT} 2020 using {M}odality
  {A}gnostic {M}eta-{L}earning}.
\newblock In \emph{Proceedings of the 17th International Conference on Spoken
  Language Translation}, pages 73--79, Online. Association for Computational
  Linguistics.

\bibitem[{Lison and Tiedemann(2016)}]{lison2016opensubtitles2016}
Pierre Lison and J{\"o}rg Tiedemann. 2016.
\newblock \href {https://www.aclweb.org/anthology/L16-1147}
  {{O}pen{S}ubtitles2016: {E}xtracting {L}arge {P}arallel {C}orpora from
  {M}ovie and {TV} {S}ubtitles}.
\newblock In \emph{Proceedings of the Tenth International Conference on
  Language Resources and Evaluation ({LREC}'16)}, pages 923--929,
  Portoro{\v{z}}, Slovenia. European Language Resources Association (ELRA).

\bibitem[{Liu et~al.(2005)Liu, Stolcke, Shriberg, and
  Harper}]{liu-etal-2005-using}
Yang Liu, Andreas Stolcke, Elizabeth Shriberg, and Mary Harper. 2005.
\newblock \href {https://doi.org/10.3115/1219840.1219896} {Using {C}onditional
  {R}andom {F}ields for {S}entence {B}oundary {D}etection in {S}peech}.
\newblock In \emph{Proceedings of the 43rd Annual Meeting of the Association
  for Computational Linguistics ({ACL}{'}05)}, pages 451--458, Ann Arbor,
  Michigan. Association for Computational Linguistics.

\bibitem[{Lu and Ng(2010)}]{lu-ng-2010-better}
Wei Lu and Hwee~Tou Ng. 2010.
\newblock \href {https://www.aclweb.org/anthology/D10-1018} {Better
  {P}unctuation {P}rediction with {D}ynamic {C}onditional {R}andom {F}ields}.
\newblock In \emph{Proceedings of the 2010 Conference on Empirical Methods in
  Natural Language Processing}, pages 177--186, Cambridge, MA. Association for
  Computational Linguistics.

\bibitem[{Matusov et~al.(2007)Matusov, Hillard, Magimai-doss, Hakkani-tur,
  Ostendorf, and Ney}]{Matusov07improvingspeech}
Evgeny Matusov, Dustin Hillard, Mathew Magimai-doss, Dilek Hakkani-tur, Mari
  Ostendorf, and Hermann Ney. 2007.
\newblock \href
  {https://isca-speech.org/archive/archive_papers/interspeech_2007/i07_2449.pdf}
  {Improving {S}peech {T}ranslation with {A}utomatic {B}oundary {P}rediction}.
\newblock In \emph{Proceedings of Interspeech}, pages 2449--2452.

\bibitem[{Mikolov et~al.(2018)Mikolov, Grave, Bojanowski, Puhrsch, and
  Joulin}]{mikolov2018advances}
Tomas Mikolov, Edouard Grave, Piotr Bojanowski, Christian Puhrsch, and Armand
  Joulin. 2018.
\newblock \href {https://www.aclweb.org/anthology/L18-1008} {Advances in
  {P}re-{T}raining {D}istributed {W}ord {R}epresentations}.
\newblock In \emph{Proceedings of the Eleventh International Conference on
  Language Resources and Evaluation ({LREC} 2018)}, Miyazaki, Japan. European
  Language Resources Association (ELRA).

\bibitem[{NIST(2017)}]{aqwv}
NIST. 2017.
\newblock \href
  {https://www.nist.gov/system/files/documents/2017/10/26/aqwv_derivation.pdf}
  {\emph{The Official Original Derivation of AQWV}}.

\bibitem[{Niu et~al.(2018)Niu, Denkowski, and Carpuat}]{niu-etal-2018-bi}
Xing Niu, Michael Denkowski, and Marine Carpuat. 2018.
\newblock \href {https://doi.org/10.18653/v1/W18-2710} {Bi-{D}irectional
  {N}eural {M}achine {T}ranslation with {S}ynthetic {P}arallel {D}ata}.
\newblock In \emph{Proceedings of the 2nd Workshop on Neural Machine
  Translation and Generation}, pages 84--91, Melbourne, Australia. Association
  for Computational Linguistics.

\bibitem[{Oard et~al.(2019)Oard, Carpuat, Galuscakova, Barrow, Nair, Niu,
  Shing, Xu, Zotkina, McKeown, Muresan, Kayi, Eskander, Kedzie, Virin, Radev,
  Zhang, Gales, Ragni, and Heafield}]{5day}
{Douglas W.} Oard, Marine Carpuat, Petra Galuscakova, Joseph Barrow, Suraj
  Nair, Xing Niu, Han-Chin Shing, Weijia Xu, Elena Zotkina, Kathleen McKeown,
  Smaranda Muresan, {Efsun Selin} Kayi, Ramy Eskander, Chris Kedzie, Yan Virin,
  {Dragomir R.} Radev, Rui Zhang, {Mark J F} Gales, Anton Ragni, and Kenneth
  Heafield. 2019.
\newblock \href {https://terpconnect.umd.edu/~oard/pdf/evia19.pdf} {Surprise
  languages: {R}apid-{R}esponse {C}ross-{L}anguage {IR}}.
\newblock In \emph{Proceedings of the Ninth International Workshop on
  Evaluating Information Access (EVIA 2019)}, pages 23--27. National Institute
  of Informatics.
\newblock The Ninth International Workshop on Evaluating Information Access
  (EVIA 2019) : a Satellite Workshop of the NTCIR-14 Conference, EVIA 2019 ;
  Conference date: 10-06-2019 Through 10-06-2019.

\bibitem[{Och(2003)}]{och-2003-minimum}
Franz~Josef Och. 2003.
\newblock \href {https://doi.org/10.3115/1075096.1075117} {Minimum {E}rror
  {R}ate {T}raining in {S}tatistical {M}achine {T}ranslation}.
\newblock In \emph{Proceedings of the 41st Annual Meeting of the Association
  for Computational Linguistics}, pages 160--167, Sapporo, Japan. Association
  for Computational Linguistics.

\bibitem[{Papineni et~al.(2002)Papineni, Roukos, Ward, and
  Zhu}]{papineni2002bleu}
Kishore Papineni, Salim Roukos, Todd Ward, and Wei-Jing Zhu. 2002.
\newblock \href {https://doi.org/10.3115/1073083.1073135} {{BLEU}: a {M}ethod
  for {A}utomatic {E}valuation of {M}achine {T}ranslation}.
\newblock In \emph{Proceedings of the 40th Annual Meeting of the Association
  for Computational Linguistics}, pages 311--318, Philadelphia, Pennsylvania,
  USA. Association for Computational Linguistics.

\bibitem[{Pham et~al.(2019)Pham, Nguyen, Ha, Hussain, Schneider, Niehues,
  St{\"u}ker, and Waibel}]{kit2019}
Ngoc-Quan Pham, Thai-Son Nguyen, Thanh-Le Ha, Juan Hussain, Felix Schneider,
  Jan Niehues, Sebastian St{\"u}ker, and Alexander Waibel. 2019.
\newblock \href {http://isl.anthropomatik.kit.edu/pdf/Pham2019b.pdf} {The
  {IWSLT} 2019 {KIT} {S}peech {T}ranslation {S}ystem}.
\newblock In \emph{International Workshop on Spoken Language Translation
  (IWSLT)}.

\bibitem[{Post(2018)}]{post-2018-call}
Matt Post. 2018.
\newblock \href {https://doi.org/10.18653/v1/W18-6319} {A {C}all for {C}larity
  in {R}eporting {BLEU} {S}cores}.
\newblock In \emph{Proceedings of the Third Conference on Machine Translation:
  Research Papers}, pages 186--191, Brussels, Belgium. Association for
  Computational Linguistics.

\bibitem[{Ragni and Gales(2018)}]{ragni18inthewild}
Anton Ragni and Mark Gales. 2018.
\newblock \href {https://doi.org/10.21437/Interspeech.2018-1085} {Automatic
  {S}peech {R}ecognition {S}ystem {D}evelopment in the ``{W}ild''}.
\newblock In \emph{Proceedings of Interspeech 2018}, pages 2217--2221.

\bibitem[{Rangarajan~Sridhar et~al.(2013)Rangarajan~Sridhar, Chen, Bangalore,
  Ljolje, and Chengalvarayan}]{Sridhar2013SegmentationSF}
Vivek~Kumar Rangarajan~Sridhar, John Chen, Srinivas Bangalore, Andrej Ljolje,
  and Rathinavelu Chengalvarayan. 2013.
\newblock \href {https://www.aclweb.org/anthology/N13-1023} {Segmentation
  {S}trategies for {S}treaming {S}peech {T}ranslation}.
\newblock In \emph{Proceedings of the 2013 Conference of the North {A}merican
  Chapter of the Association for Computational Linguistics: Human Language
  Technologies}, pages 230--238, Atlanta, Georgia. Association for
  Computational Linguistics.

\bibitem[{Rao et~al.(2007)Rao, Lane, and Schultz}]{rao2007optimizing}
Sharath Rao, Ian Lane, and Tanja Schultz. 2007.
\newblock \href
  {https://www.csl.uni-bremen.de/cms/images/documents/publications/RaoSchultz_Interspeech2007.pdf}
  {Optimizing {S}entence {S}egmentation for {S}poken {L}anguage {T}ranslation}.
\newblock In \emph{Proceedings of Interspeech}.

\bibitem[{Rosales~N{\'u}{\~n}ez et~al.(2019)Rosales~N{\'u}{\~n}ez, Seddah, and
  Wisniewski}]{rosales-nunez-etal-2019-comparison}
Jos{\'e}~Carlos Rosales~N{\'u}{\~n}ez, Djam{\'e} Seddah, and Guillaume
  Wisniewski. 2019.
\newblock \href {https://www.aclweb.org/anthology/W19-6101} {Comparison between
  {NMT} and {PBSMT} {P}erformance for {T}ranslating {N}oisy {U}ser-{G}enerated
  {C}ontent}.
\newblock In \emph{Proceedings of the 22nd Nordic Conference on Computational
  Linguistics}, pages 2--14, Turku, Finland. Link{\"o}ping University
  Electronic Press.

\bibitem[{Sennrich et~al.(2016)Sennrich, Haddow, and
  Birch}]{sennrich-etal-2016-neural}
Rico Sennrich, Barry Haddow, and Alexandra Birch. 2016.
\newblock \href {https://doi.org/10.18653/v1/P16-1162} {Neural {M}achine
  {T}ranslation of {R}are {W}ords with {S}ubword {U}nits}.
\newblock In \emph{Proceedings of the 54th Annual Meeting of the Association
  for Computational Linguistics (Volume 1: Long Papers)}, pages 1715--1725,
  Berlin, Germany. Association for Computational Linguistics.

\bibitem[{Senter and Smith(1967)}]{ari}
R.J. Senter and E.A. Smith. 1967.
\newblock \href {https://apps.dtic.mil/dtic/tr/fulltext/u2/667273.pdf}
  {Automated {R}eadability {I}ndex}.
\newblock \emph{AMRL-TR. Aerospace Medical Research Laboratories}.

\bibitem[{Shriberg et~al.(2000)Shriberg, Stolcke, Hakkani-T\"{u}r, and
  T\"{u}r}]{10.1016/S0167-6393(00)00028-5}
Elizabeth Shriberg, Andreas Stolcke, Dilek Hakkani-T\"{u}r, and G\"{u}khan
  T\"{u}r. 2000.
\newblock \href {https://doi.org/10.1016/S0167-6393(00)00028-5}
  {Prosody-{B}ased {A}utomatic {S}egmentation of {S}peech into {S}entences and
  {T}opics}.
\newblock \emph{Speech Communication}, 32(1–2):127–154.

\bibitem[{Sperber et~al.(2019)Sperber, Neubig, Niehues, and
  Waibel}]{sperber-etal-2019-attention}
Matthias Sperber, Graham Neubig, Jan Niehues, and Alex Waibel. 2019.
\newblock \href {https://doi.org/10.1162/tacl_a_00270} {Attention-{P}assing
  {M}odels for {R}obust and {D}ata-{E}fficient {E}nd-to-{E}nd {S}peech
  {T}ranslation}.
\newblock \emph{Transactions of the Association for Computational Linguistics},
  7:313--325.

\bibitem[{Sperber et~al.(2018)Sperber, Pham, Nguyen, Niehues, M{\"u}ller, Ha,
  St{\"u}ker, and Waibel}]{kit2018}
Matthias Sperber, Ngoc~Quan Pham, Thai~Son Nguyen, Jan Niehues, Markus
  M{\"u}ller, Thanh-Le Ha, Sebastian St{\"u}ker, and Alexander Waibel. 2018.
\newblock \href {https://isl.anthropomatik.kit.edu/downloads/Sperber2018a.pdf}
  {{KIT}’s {IWSLT} 2018 {SLT} {T}ranslation {S}ystem}.
\newblock In \emph{International Workshop on Spoken Language Translation
  (IWSLT)}.

\bibitem[{Stolcke and Shriberg(1996)}]{Stolcke1996AutomaticLS}
Andreas Stolcke and Elizabeth Shriberg. 1996.
\newblock \href {http://www.asel.udel.edu/icslp/cdrom/vol2/715/a715.pdf}
  {Automatic {L}inguistic {S}egmentation of {C}onversational {S}peech}.
\newblock \emph{Proceeding of Fourth International Conference on Spoken
  Language Processing. ICSLP '96}, 2:1005--1008 vol.2.

\bibitem[{Strohman et~al.(2005)Strohman, Metzler, Turtle, and
  Croft}]{strohman2005indri}
Trevor Strohman, Donald Metzler, Howard Turtle, and W.~Croft. 2005.
\newblock \href {http://ciir.cs.umass.edu/pubfiles/ir-407.pdf} {Indri: {A}
  language-model based search engine for complex queries}.
\newblock \emph{Information Retrieval}.

\bibitem[{Vaswani et~al.(2017)Vaswani, Shazeer, Parmar, Uszkoreit, Jones,
  Gomez, Kaiser, and Polosukhin}]{NIPS2017_7181}
Ashish Vaswani, Noam Shazeer, Niki Parmar, Jakob Uszkoreit, Llion Jones,
  Aidan~N Gomez, {\L}ukasz Kaiser, and Illia Polosukhin. 2017.
\newblock \href
  {http://papers.nips.cc/paper/7181-attention-is-all-you-need.pdf} {Attention
  is {A}ll {Y}ou {N}eed}.
\newblock In I.~Guyon, U.~V. Luxburg, S.~Bengio, H.~Wallach, R.~Fergus,
  S.~Vishwanathan, and R.~Garnett, editors, \emph{Advances in Neural
  Information Processing Systems 30}, pages 5998--6008. Curran Associates, Inc.

\bibitem[{Wan et~al.(2020)Wan, Jiang, Kedzie, Turcan, Bell, and
  McKeown}]{wan2020subtitles}
David Wan, Zhengping Jiang, Chris Kedzie, Elsbeth Turcan, Peter Bell, and Kathy
  McKeown. 2020.
\newblock \href {https://www.aclweb.org/anthology/2020.clssts-1.11} {Subtitles
  to {S}egmentation: {I}mproving {L}ow-{R}esource {S}peech-to-{T}ext
  {T}ranslation {P}ipelines}.
\newblock In \emph{Proceedings of the workshop on Cross-Language Search and
  Summarization of Text and Speech (CLSSTS2020)}, pages 68--73, Marseille,
  France. European Language Resources Association.

\bibitem[{Weiss et~al.(2017)Weiss, Chorowski, Jaitly, Wu, and
  Chen}]{Weiss2017SequencetoSequenceMC}
Ron~J. Weiss, Jan Chorowski, Navdeep Jaitly, Yonghui Wu, and Zhifeng Chen.
  2017.
\newblock \href {https://doi.org/10.21437/Interspeech.2017-503}
  {Sequence-to-sequence {M}odels {C}an {D}irectly {T}ranslate {F}oreign
  {S}peech}.
\newblock In \emph{Proceedings of Interspeech 2017}, pages 2625--2629.

\bibitem[{Welch(1947)}]{10.1093/biomet/34.1-2.28}
B.~L. Welch. 1947.
\newblock \href {https://doi.org/10.1093/biomet/34.1-2.28} {{THE GENERALIZATION
  OF ‘STUDENT'S’ PROBLEM WHEN SEVERAL DIFFERENT POPULATION VARLANCES ARE
  INVOLVED}}.
\newblock \emph{Biometrika}, 34(1-2):28--35.

\bibitem[{Zhang et~al.(2020)Zhang, Karakos, Hartmann, Srivastava, Tarlin,
  Akodes, Gouda, Bathool, Zhao, Jiang, Schwartz, and
  Makhoul}]{zhang-etal-2020-2019}
Le~Zhang, Damianos Karakos, William Hartmann, Manaj Srivastava, Lee Tarlin,
  David Akodes, Sanjay~Krishna Gouda, Numra Bathool, Lingjun Zhao, Zhuolin
  Jiang, Richard Schwartz, and John Makhoul. 2020.
\newblock \href {https://www.aclweb.org/anthology/2020.clssts-1.8} {The 2019
  {BBN} {C}ross-lingual {I}nformation {R}etrieval {S}ystem}.
\newblock In \emph{Proceedings of the workshop on Cross-Language Search and
  Summarization of Text and Speech (CLSSTS2020)}, pages 44--51, Marseille,
  France. European Language Resources Association.

\end{thebibliography}
\bibliographystyle{acl_natbib}

\clearpage
\appendix

\section{Full Complexity Analysis}
\label{complexity}

We present the full results of our complexity analysis as described in \autoref{sec:complexity}. Bulgarian (\autoref{tab:bleu_quartiles_bg_full}, Lithuanian (\autoref{tab:bleu_quartiles_lt_full}), and Farsi (\autoref{tab:bleu_quartiles_fa_full}) results are shown for all three MT models as well as both the acoustic segmentation and our \textit{Seg + Sub} segmentation correction model. The best score for each MT system and quartile is bolded.

\vfill

\begin{table}[h!]
\centering
\begin{tabular}{ ll ccc}
\toprule
 ARI & \multirow{2}{*}{Model} & EDI & UMD & UMD \\
  Quartile & & NMT & NMT & SMT \\
\midrule
\multirow{2}{*}{Q1} & Acous. & 17.78 & \textbf{24.70} & 22.26 \\
& Seg + Sub & \textbf{17.92} & 11.62 & \textbf{22.58} \\
\midrule
\multirow{2}{*}{Q2} & Acous.  & 20.76 & 22.09 & 23.35 \\
& Seg + Sub & \textbf{26.89} & \textbf{31.24} & \textbf{23.43} \\
\midrule
\multirow{2}{*}{Q3} & Acous. & 22.87 & 20.38 & \textbf{22.45} \\
& Seg + Sub & \textbf{29.96} & \textbf{33.22} & 21.68 \\
\midrule
\multirow{2}{*}{Q4} & Acous.  & 23.41 & 21.00 & 22.97 \\
& Seg + Sub & \textbf{29.87} & \textbf{35.24} & \textbf{23.25} \\
\bottomrule
\end{tabular}
\caption {Bulgarian BLEU scores on Test (Small) (transcribed portion) when separated into quartiles by sentence complexity (as measured by ARI).}
\label{tab:bleu_quartiles_bg_full}
\end{table}
\begin{table}[h!]
\centering
\begin{tabular}{ ll ccc}
\toprule
 ARI & \multirow{2}{*}{Model} & EDI & UMD & UMD \\
  Quartile & & NMT & NMT & SMT \\
\midrule
\multirow{2}{*}{Q1} & Acous. & 4.24 & \textbf{3.13} & \textbf{5.14} \\
& Seg + Sub & \textbf{4.32} & 2.76 & 5.11 \\
\midrule
\multirow{2}{*}{Q2} & Acous.  & \textbf{14.07} & \textbf{7.18} & \textbf{13.45} \\
& Seg + Sub & 14.01 & 6.83 & 13.39 \\
\midrule
\multirow{2}{*}{Q3} & Acous. & 15.95 & 7.81 & 14.85 \\
& Seg + Sub & \textbf{16.39} & \textbf{8.13} & \textbf{14.87} \\
\midrule
\multirow{2}{*}{Q4} & Acous.  & \textbf{15.41} & \textbf{7.94} & 15.85 \\
& Seg + Sub & 14.60 & 7.78 & \textbf{15.87} \\
\bottomrule
\end{tabular}
\caption {Lithuanian BLEU scores on Test (Small) (transcribed portion) when separated into quartiles by sentence complexity (as measured by ARI).}
\label{tab:bleu_quartiles_lt_full}
\end{table}
\begin{table}[ht!]
\centering
\begin{tabular}{ ll ccc}
\toprule
 ARI & \multirow{2}{*}{Model} & EDI & UMD & UMD \\
  Quartile & & NMT & NMT & SMT \\
\midrule
\multirow{2}{*}{Q1} & Acous.  & 3.46 & 3.40 & 3.21 \\
& Seg + Sub & \textbf{6.05} & \textbf{4.49} & \textbf{3.43} \\
\midrule
\multirow{2}{*}{Q2} & Acous. & 3.55 & 3.87 & 3.22 \\
& Seg + Sub & \textbf{5.06} & \textbf{5.41} & \textbf{3.66} \\
\midrule
\multirow{2}{*}{Q3} & Acous. & 4.71 & 5.33 & 5.19 \\
& Seg + Sub & \textbf{7.44} & \textbf{7.29} & \textbf{5.54} \\
\midrule
\multirow{2}{*}{Q4} & Acous.  & 5.88 & 6.15 & 3.90 \\
& Seg + Sub & \textbf{7.81} & \textbf{7.59} & \textbf{4.47} \\
\bottomrule
\end{tabular}
\caption {Farsi BLEU scores on Test (Small) (transcribed portion) when separated into quartiles by sentence complexity (as measured by ARI).}
\label{tab:bleu_quartiles_fa_full}
\end{table}

\end{document}